\documentclass[letterpaper, 10 pt, conference]{ieeeconf}  

\IEEEoverridecommandlockouts                              

\newcommand{\href}[1]{#1} 

\usepackage{amsmath,amssymb,amstext} 
\usepackage[pdftex]{graphicx} 
\usepackage{booktabs}
\usepackage{multirow}
\usepackage{subcaption}
\graphicspath{ {./images/} }
\usepackage{balance}

\usepackage[pdftex,pagebackref=false]{hyperref} 
\usepackage{orcidlink}
\usepackage{cleveref}
\usepackage{lipsum}

\title{\LARGE \bf
A Hierarchical Pedestrian Behavior Model to Generate\\Realistic Human Behavior in Traffic Simulation
}

\author{Scott~Larter\,\authorrefmark{2},~Rodrigo~Queiroz,~Sean~Sedwards\,\orcidlink{0000-0002-2903-0823},~Atrisha~Sarkar, and~Krzysztof~Czarnecki\,\orcidlink{0000-0003-1642-1101} 
\thanks{
This work was supported by Natural Sciences and Engineering Research Council of Canada (NSERC), Discovery Grant RGPIN-2017-04733.
Author SS was supported by Japanese
Science and Technology agency (JST) ERATO project JPMJER1603: HASUO Metamathematics for Systems Design.}
\thanks{
\authorrefmark{2}\,Corresponding author. All authors are with the University of Waterloo, 200 University Avenue West, Waterloo, Ontario, Canada:\newline
{\tt\scriptsize \{scott.larter, rodrigo.queiroz, sean.sedwards, atrisha.sarkar, krzysztof.czarnecki\}@uwaterloo.ca}}
}

\begin{document}

\maketitle
\thispagestyle{empty}
\pagestyle{empty}

\begin{abstract}

Modelling pedestrian behavior is crucial in the development and testing of autonomous vehicles. In this work, we present a hierarchical pedestrian behavior model that generates high-level decisions through the use of behavior trees, in order to produce maneuvers executed by a low-level motion planner using an adapted Social Force model. A full implementation of our work is integrated into GeoScenario Server, a scenario definition and execution engine, extending its vehicle simulation capabilities with pedestrian simulation. The extended environment allows simulating test scenarios involving both vehicles and pedestrians to assist in the scenario-based testing process of autonomous vehicles. The presented hierarchical model is evaluated on two real-world data sets collected at separate locations with different road structures. Our model is shown to replicate the real-world pedestrians' trajectories with a high degree of fidelity and a decision-making accuracy of 98\% or better, given only high-level routing information for each pedestrian.

\end{abstract}


\section{Introduction} \label{sec:intro}

Modelling pedestrian behavior in traffic environments is a crucial step in the development and testing of autonomous vehicles (AV) and automated driving systems (ADS). As the environment's most vulnerable road users, misinterpretation of their behavior by an AV can lead to catastrophic consequences. Through rigorous testing of a wide range of traffic scenarios, safe interactions between AVs and pedestrians can be ascertained. A realistic and highly controllable pedestrian simulation model that supports the scenario-based testing process is a valuable tool in testing AV capabilities and responses to critical situations.

In this work, we present a hierarchical pedestrian behavior model that incorporates behavior trees to handle high-level decision-making processes and an adapted Social Force Model to drive low-level motion. Our model caters to scenario-based testing \cite{Kentaro2004} as it provides an explicit representation of decision processes, allowing engineers to inject desired pedestrian behaviors into existing scenarios. Through this process, rare and critical situations, which may be scarce or absent in existing data, can be generated and tested on AV systems to evaluate their responses safely in a simulation environment.

\begin{figure}[t]
    \centering
    \vspace{1em}
    \begin{subfigure}{0.48\linewidth}
    \includegraphics[width=\linewidth]{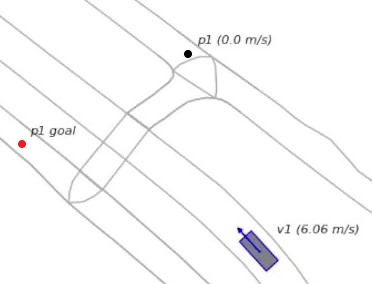}
    \caption{Pedestrian (p1) speed is 0\,m/s as vehicle's collision with crosswalk is not yet detected.}
    \label{subfig:test_scenario_frame_1}
    \end{subfigure}
    ~
    \begin{subfigure}{0.48\linewidth}
    \includegraphics[width=\linewidth]{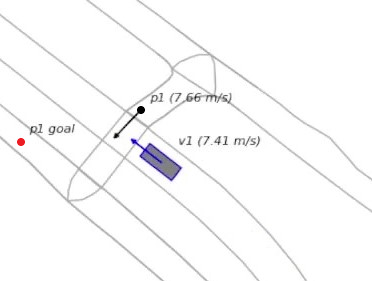}
    \caption{p1 dynamically adjusts speed to ensure collision with v1 at crosswalk.}
    \label{subfig:test_scenario_frame_2}
    \end{subfigure}
    \caption{Simulation of scenario in which a collision is guaranteed between a pedestrian and a vehicle at a crosswalk}
    \label{fig:test_scenario_ex}
\end{figure}

For example, consider the situation of a pedestrian running into the roadway directly in front of a vehicle, causing a collision. If the goal of this test scenario is to cause a collision independent of the agents' starting positions and velocities, recreation of this scenario is difficult with a simple trajectory-based pedestrian that only follows a set speed profile. However, with our model, we can dynamically adjust the pedestrian's speed to ensure a collision occurs, independent of the vehicle's approaching distance and speed. Pedestrians with pre-defined trajectories would require constant manual adjustments in their positioning and speed profiles whenever changes are made to the vehicle's approaching parameters. \Cref{fig:test_scenario_ex} visualizes our model recreating this situation in a test simulation scenario.

Within the context of traffic simulation, existing microscopic models have a heavy focus on specific interaction scenarios \cite{Millard-Ball2018} \cite{CamaraFanta2018} \cite{LiChuanyao2021} leaving little room for extensibility or they employ a ``black-box'' or non-deterministic approach \cite{Li2012} \cite{Feng2013} \cite{Suo2021} where the pedestrian's actions and overall trajectory can not be derived given the scenario set up and inputs. There are existing models, including numerous Social Force and cellular automata models \cite{Helbing1998} \cite{Blue2001} \cite{Zhang2012}, that demonstrate higher-level behaviors and thought-processes, such as lane formation and bottle-necking; however, they largely do not allow engineers to customize these behaviors or inject specific desired actions into a test scenario. The presented simulation model provides a high degree of control over pedestrian actions and behaviors currently lacking in literature. Our model is explicitly decision-driven, through the use of behavior trees, while still producing realistic low-level movements. Such functionality can greatly benefit the scenario-based testing of AVs, since testers can force rare or possibly dangerous situations involving pedestrians in a safe simulation environment.

Aside from singular-concept approaches, multi-layered models have been developed to address pedestrian behaviors in traffic. Conjunctive use of trajectory-planning, rule-based, and Social Force layers have been implementing to assist the design of shared spaces \cite{Anvari2015} \cite{Rinke2017}. Another common approach is to combine a popular microscopic model, such as a cellular automaton, with higher-level game theoretic concepts \cite{Tanimoto2010} \cite{Guan2016} \cite{Lin2018} \cite{Wu2019}, to construct a multi-layered model in which high and low-level interactions can be handled by separate components. As before, such models lack the flexibility provided by behavior trees, which our model exposes to test engineers as a domain-specific language and allows them to create and customize diverse and realistic scenarios.

In summary, our work makes the following contributions:
\begin{enumerate}
    \item a novel model for pedestrian motion simulation, with (i) highly controllable and customizable decision making via behavior trees, including a catalog of reusable pedestrian maneuvers and decision conditions, and (ii) realistic motion via an adapted Social Force Model;
    \item an evaluation of  the decision-making realism and motion fidelity of the model on two traffic trajectory data sets, showing the ability to replicate (i) the high-level decisions of the empirical pedestrians with a 98\% or better accuracy and (ii) the motion trajectories with an average deviation of 1.36\,m.
\end{enumerate}


\section{Background} \label{sec:background}
In this section we briefly describe some of the background technology and concepts used in our approach.

\subsection{Social Force Model} \label{subsec:sfm}

The Social Force model (SFM) treats pedestrian motion as if it is based on attractive and repulsive physical forces. Our model incorporates a variation of the classical SFM~\cite{Helbing2000} with additional forces. In its classical version, the SFM applies three main collections of forces to the agent: an attracting force drawing the agent towards their destination, a repelling force from each of the other agents in the scene, and a repelling force from each wall or border in the environment. These three forces are summed into an acceleration equation that describes the pedestrian's change in velocity throughout the scenario's duration.

The attracting force, \(\textbf{f}_\text{adapt}\), is responsible for propelling pedestrian \(i\) in the direction \(\textbf{e}_i^0(t)\), pointing towards their destination point. Given a current velocity, \(\textbf{v}_i(t)\), and a desired speed, \(v_i^0(t)\), the attracting force acting on pedestrian \(i\) with mass \(m_i\) is

\[ \textbf{f}_\text{adapt} = m_i \frac{v_i^0(t)\textbf{e}_i^0(t) - \textbf{v}_i(t)}{\tau_i} \]

\noindent
where their acceleration is over characteristic time \(\tau_i\).

Repelling forces acting on the pedestrian are divided into three groups of forces: \(\textbf{f}_\text{otherPeds}\), \(\textbf{f}_\text{vehicles}\), and \(\textbf{f}_\text{borders}\). Each of these forces is composed of a sum of sub-forces directed away from an object (pedestrian, vehicle, or wall). They describe the human tendency to avoid collision with and maintain a comfortable distance from other objects and agents in their surroundings. Further details on the \(\textbf{f}_\text{otherPeds}\) and \(\textbf{f}_\text{borders}\) forces can be found in Helbing et al.'s model \cite{Helbing2000} and \(\textbf{f}_\text{vehicles}\) is presented by Anvari et al.\ \cite{Anvari2015}.

The resulting sum of forces that drives each pedestrian's motion is given by

\begin{equation}
    \label{eq:sfm}
    m_i \, \frac{d\textbf{v}_i}{dt} = \textbf{f}_\text{adapt} + \textbf{f}_\text{otherPeds} + \textbf{f}_\text{vehicles} + \textbf{f}_\text{borders}
\end{equation}

\subsection{Behavior Trees} \label{BT_background}

Behavior trees can be used to concisely and explicitly model a wide range of decision making processes~\cite{Colledanchise2018}. Within our model, behavior trees are used to select an appropriate maneuver for each pedestrian at each simulation cycle. Each pedestrian contains a personal behavior tree which, at each time step in a given scenario, is ``ticked''. The ticking process traverses the tree with a certain path and ultimately outputs a selected maneuver. The behavior trees used in the presented model are composed of four types of nodes: \emph{selector}, \emph{sequence}, \emph{maneuver}, and \emph{condition}. Selectors and sequences are internal nodes and control the path of the tree traversal, while maneuvers and conditions are the tree's leaf nodes. Each leaf node, after being evaluated, returns a status to their parent node. Valid statuses are \textbf{Success}, \textbf{Failure}, and \textbf{Running}. These statuses affect which nodes are visited next by the tick.

Selector nodes, denoted by a question mark (?), are analogous to a short-circuit OR in that they tick their child nodes sequentially from left to right until a status of \textbf{Success} or \textbf{Running} is received. This status is then returned to the selector's parent node. If all of the child nodes return \textbf{Failure}, then \textbf{Failure} is returned to the selector's parent. On the other hand, sequence nodes, denoted by an arrow (\(\rightarrow\)), are similar to a short-circuit AND. They tick their child nodes sequentially from left to right until a status of \textbf{Failure} is received (or they run out of child nodes). Maneuvers and conditions are leaf nodes that are evaluated and return one of the three statuses to their internal parent node. The maneuver node that is visited last before the tick returns from the entire tree becomes the selected maneuver.

\begin{figure}[ht]
    \centering
    \includegraphics[width=.75\linewidth]{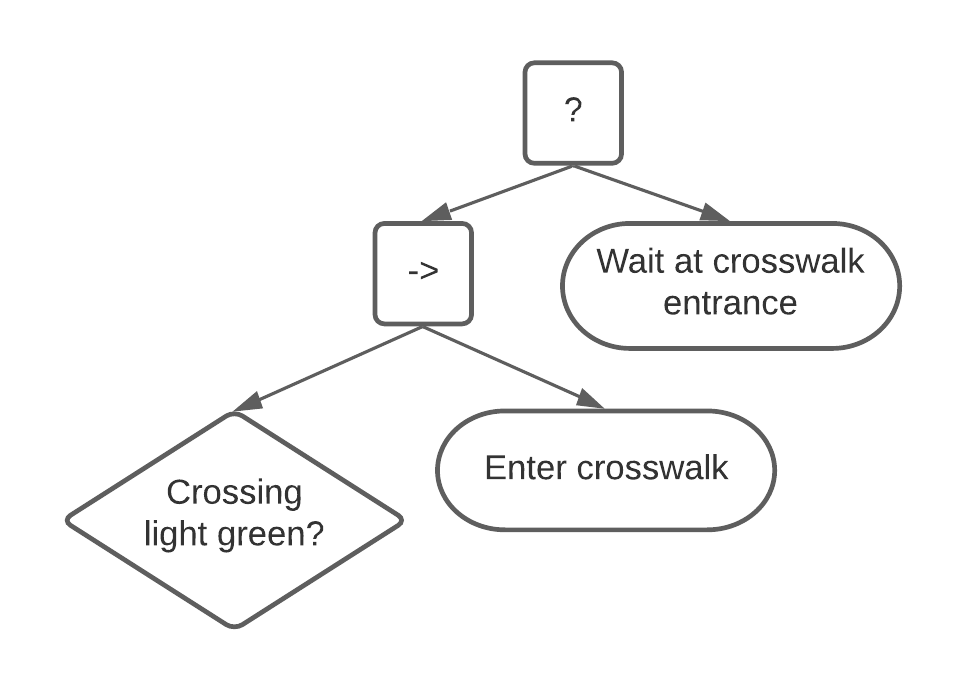}
    \caption{Simple behavior tree for entering crosswalk decision demonstrating the use of condition (diamond), maneuver (ellipse), selector (?), and sequence (\(\rightarrow\)) nodes.}
    \label{fig:btree_example}
\end{figure}

As a simple example, \Cref{fig:btree_example} shows a valid behavior tree an agent could use to determine whether to enter a crosswalk. The process flow of this tree first checks if the crossing signal is green. If so, the pedestrian enters the crosswalk and waits at the entrance otherwise. In the case that the crossing signal is not green, the tick returns from the left side of the tree with a status of \textbf{Failure} (from the diamond condition node). Since the root is a \emph{selector} node, it proceeds to tick the right side, consisting solely of a single maneuver node which will return a non-failure status and become the selected maneuver.

\section{Model Design} \label{sec:model-design}

To introduce our model, we first discuss the design requirements necessary for a practical pedestrian simulation model.

\subsection{Facilitation of Scenario Creation}

An essential component of scenario-based testing is a straight-forward and clear scenario creation process~\cite{modes2020-01-1204}. Test scenarios involving pedestrians running our model are expressed in GeoScenario~\cite{Queiroz2019}. GeoScenario is a domain-specific language for simple and extensible representation of traffic scenarios, built using the Open Street Map (OSM) standard. All of the essential components of a scenario can be represented in GeoScenario, including dynamic agents, static objects, and the underlying Lanelet2 map, among others. Through GeoScenario, pedestrian agents are represented by nodes containing tags to describe their individual attributes. As GeoScenario abides by the Open Street Map standard, we employ the OSM scenario editor tool, JOSM\footnote{\url{https://josm.openstreetmap.de}}, to create and edit scenarios. JOSM provides engineers with a simple visual tool to quickly add, move, and tag nodes and ways. Pedestrian-specific agents must include tags defining their identifying name, destination point, and the file containing their personal behavior tree.

\subsection{Customizable Pedestrian Behavior}

Behavior trees are an integral part of each scenario as they provide the tester with a fine-level of control over how each pedestrian behaves in a given context. Each pedestrian is assigned a behavior tree file in the scenario definition file which dictates their decision process at each simulation cycle. Our model comes with a library of reusable behavior trees that represent common behaviors, such as walking along a sidewalk and traversing signalized and unsignalized crosswalks. The explicit nature of behavior trees in representing discrete decision-making processes lends itself well to forcing desired behaviors within scenarios. Testers can intentionally trigger particular actions that can lead to critical or dangerous situations; for example, pedestrians unexpectedly running in front of a moving vehicle. The behavior trees designed for our model are also augmented with a number of tunable parameters that define the different ways pedestrians may execute the same maneuver. When importing behavior trees from a library, their parameters and sub-trees can be overridden to define the desired behavior.

\subsection{Dynamic Interactions Between Agents}

Dynamic interactions between agents within a scenario are handled by both the Social Force model and behavior tree components of our model, though in different ways. The Social Force model (SFM) is responsible for immediate, reactionary interactions with vehicles and other pedestrians. At the trajectory level, a collection of repulsive forces is applied to the pedestrian when another agent is in its proximity that causes the pedestrian to naturally avoid collisions. Behavior trees are designed to handle higher-level interactions proactively. Depending on their composition, behavior trees can check the relative states of other agents, such as their distance and speed, to output an informed response to another agent's actions. If required by a scenario, behavior trees can selectively and conditionally modify the forces for specific pedestrians by manipulating SFM parameters.

\subsection{Realistic Human Movements and Decisions}

A practical pedestrian behavior model needs to be able to simulate realistic movements and rational decision-making processes. We evaluate this requirement on our model against two naturalistic data sets with different road structures. Our hierarchical approach of handling high-level decisions with customizable behavior trees that inform low-level trajectory movements is shown to be effective in producing realistic movements and decisions that map to real-world scenarios.

\section{Model Architecture}

\begin{figure}[ht]
    \centering
    \vspace{0.5em}
    \includegraphics[width=\linewidth]{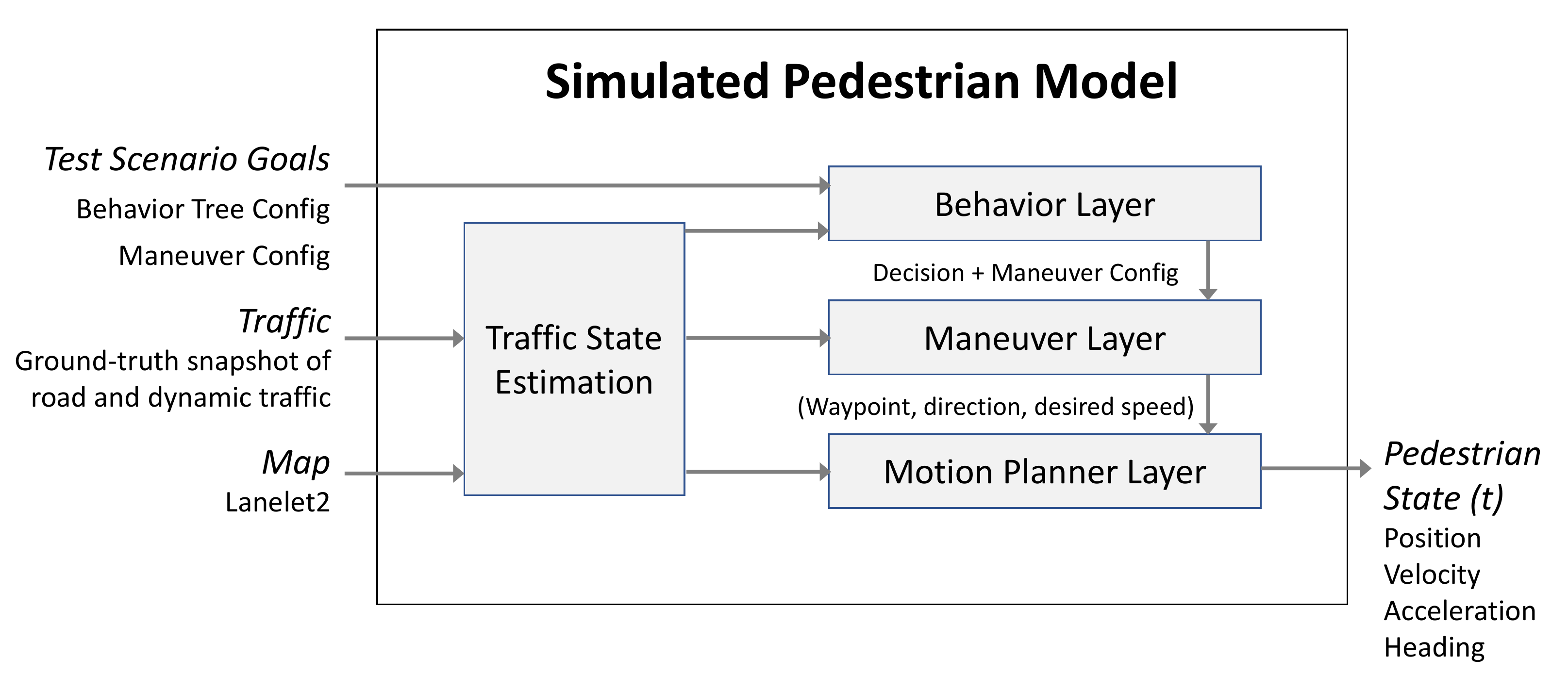}
    \caption{Multi-layered model diagram with process and information flow}
    \label{fig:model_process_flow}
\end{figure}

The model structure is composed of three layers: the Behavior layer, Maneuver layer, and Motion Planner layer. As an overview, the Behavior layer receives the environment state representation and decides on an appropriate maneuver to execute. This maneuver is then passed to the Maneuver layer, which plans how best to execute the selected maneuver. The layer forms instructions on how to adapt the current trajectory in the form of a vector containing the pedestrian's updated waypoint, direction vector, and desired speed, to pass to the Motion Planner layer. When the Motion Planner layer receives these instructions, it feeds the passed vector into the Social Force Model to determine the state information of the pedestrian for the next time step. Finally, this state information is updated and reflected in the environment. This process flow is visualized in~\Cref{fig:model_process_flow}.

The simulation traffic environment is represented in the two dimensional Cartesian coordinate frame and all computations for pedestrian movements are calculated within this frame. The physical traffic structure of the world is represented by Lanelet2~\cite{Poggenhans2018} compatible map files. Dynamic elements of the environment include the changing states of vehicles, other pedestrians, and traffic lights for both vehicle lanes and crosswalks.

We define two types of pedestrian agents: Empirical Pedestrians (EP) and Simulated Pedestrians (SP). Empirical Pedestrians follow a predefined sequence of trajectory points with preset time intervals and do not use our model to drive their motion. Simulated pedestrians do apply the presented model to determine their movements and are the focus of this paper. Agents' state information contains two dimensional positional, velocity, and acceleration components as well as heading in the vector \(|(x, \dot{x}, \ddot{x}, y, \dot{y}, \ddot{y}, \theta)|_t\). In our evaluation, Empirical Vehicle (EV) agents are used in the experiments; however, our pedestrian model does not depend on the particular implementation of vehicle agents, which allows for targeted testing of a desired ADS or AV system. The state vector format is shared between all types of pedestrian and vehicle agents.

\subsection{Behavior Layer}

The Behavior layer is tasked with determining an appropriate maneuver to execute given the current environmental context. At scenario creation, each SP agent is assigned a text-based behavior tree file representing their personal tree. The Behavior layer consumes the current traffic state estimation as well as configuration parameters for the overall tree and each possible maneuver. The output of this layer is a selected maneuver with its specific configuration parameters, which is subsequently passed to the Maneuver layer. \Cref{tab:maneuvers_and_conditions} contains the available maneuvers and conditions currently implemented in the model. Behavior trees are constructed with a subset of these lists as leaf nodes. This maneuver and condition catalog was derived manually in the process of developing the behavior trees required to simulate each of the pedestrians in the data sets discussed in Section~\ref{sec:evaluation}. 

\begin{table}[ht]
    \centering
    \caption{Maneuvers and conditions implemented in model}
    \label{tab:maneuvers_and_conditions}
    \begin{tabular}{l l}
        \toprule
        Maneuvers & Conditions \\
        \midrule
        Keep in Lane & Reached goal \\
        Stop & Has target crosswalk \\
        Enter Crosswalk & Approaching target crosswalk \\
        Wait at Crosswalk & At target crosswalk entrance \\
        Increase Walking Speed & At target crosswalk exit \\
        Select Crosswalk by Light State & Waiting at target crosswalk entrance \\
        Return to Crosswalk Entrance & Target crosswalk has signal \\
        & Crossing signal is green/red/yellow \\
        & Can cross before signal turns red \\
        & Vehicle approaching crosswalk \\
        \bottomrule
    \end{tabular}
\end{table}

As explained and visualized in Section~\ref{BT_background}, the internal nodes of a behavior tree are sequence and selector nodes and the leaf nodes are maneuver, condition, and, optionally, sub-tree nodes. The internal nodes dictate the traversal path of the ``tick'' based on the return statuses of leaf nodes. A sub-tree is itself a complete behavior tree that can be ``plugged in'' in place of a leaf node of a different tree. Sub-trees are simply an extension of their base tree that are useful to maintain modularity and, in our case, add varying levels of behavior to control how different pedestrians react to the same situation. In our model, we have designed three sub-trees, representing three levels of aggressiveness, to define different decision-making processes when a pedestrian is planning to enter a crosswalk. We label the three levels of aggressiveness as Low, Medium, and High. Briefly, low aggressiveness pedestrians only enter the crosswalk when the crossing signal is green (or there is no crossing signal at all). A medium aggressiveness pedestrian enters on green, but also on a yellow signal when they judge that they can sufficiently cross before the red signal. Finally, the pedestrians with a high level of aggressiveness choose to enter the crosswalk regardless of the signal state as long as they are not put in danger by a vehicle in doing so.

At each simulation cycle, the behavior tree of each pedestrian is ticked and an appropriate maneuver is selected to be executed by the following layers. This maneuver is passed as input to the Maneuver layer for interpretation.

\subsection{Maneuver Layer}

The job of the Maneuver layer is to translate a maneuver received from the Behavior layer into instructions on how the pedestrian should adjust their trajectory. The instructions must be interpretable by the subsequent Motion Planner layer. Each received maneuver is converted into low-level instructions containing the following three components: a waypoint, a desired direction unit vector, and a desired speed. A waypoint is defined as an intermediate goal point the pedestrian visits before its final destination. As an example of translating a selected maneuver, if the \emph{Enter Crosswalk} maneuver is received, the Maneuver layer determines the updated waypoint to be a point at the end of the crosswalk, the desired direction vector to be the unit vector pointing to the new waypoint, and the desired speed to be the default desired speed of the pedestrian. These three components are passed to the next layer, the Motion Planner layer, which handles the execution of the selected maneuver.

\subsection{Motion Planner Layer}

The Motion Planner layer is driven by an adaptation of the Social Force Model. It receives the three components passed by the Maneuver layer describing changes to the pedestrian's trajectory. This layer also has access to the traffic state estimation and therefore all positions and velocities of the other agents in the scene. Our model implements the classic SFM described in Section~\ref{subsec:sfm} extended by Anvari's method \cite{Anvari2015} for handling pedestrian-vehicle interactions. The parameters of our SFM were manually calibrated through simulation testing. Our traffic environments in Section~\ref{sec:evaluation}) did not contain any walls, and thus the wall forces are currently not included. Walking within a sidewalk or crosswalk's bounds is handled through the \emph{Keep in Lane} maneuver, which uses a desired walking direction vector (that may not necessarily point to the waypoint) to guide the pedestrian within the sidewalk or crosswalk (i.e., a pedestrian lane) and bias it along the right or left lane-boundary.
Wall forces may be added for future environments containing physical walls.

The Motion Planner layer is tasked with running an iteration of the Social Force model's formula~\eqref{eq:sfm} to determine the model pedestrian's change in velocity at each simulation time step. Subsequently, this layer directly updates the pedestrian's state information before the next simulation cycle.

\section{Implementation}

The presented model is written in Python and integrated into GeoScenario Server, a full scenario simulation environment capable of running traffic scenarios as a standalone application. The Server provides the necessary infrastructure for full scenario simulation. It parses GeoScenario scenario files and initializes the necessary elements (pedestrians and vehicle agents, agent goal points, traffic lights, etc.), reads and loads the Lanelet2 map file to provide the underlying road network structure for the scenario, maintains the static and dynamic environment and facilitates the information flow between agents and their surrounding environmental context, and finally runs the scenario by iterating through each of its agents to update their state in the environment at each simulation cycle. Vehicles are simulated with the SDV model from Queiroz et al.~\cite{Queiroz2022_arXiv}. Pedestrians originally used a simple model with Pre-defined Trajectories (PDTs). We extended the server by adding our pedestrian model as an alternative to replace the PDT-based model.

GeoScenario Server provides the optional integration with WISE Sim, a simulator based on UnrealEngine\footnote{\url{https://www.unrealengine.com}} to run the WISE Automated Driving System. The server incorporates a shared-memory interface with WISE Sim to integrate its dynamic agents and scenario environment into the simulator, but it also features an experimental integration with the CARLA simulator~\cite{CARLA-pmlr-v78-dosovitskiy17a}. More details on the GeoScenario Server and how to use the pedestrian model can be found at {\small\url{https://geoscenario2.readthedocs.io}}.



\section{Evaluation} \label{sec:evaluation}

We assess our model in terms of how well it can reproduce low-level trajectories and high-level decisions observed in a naturalistic data set, as well as its extensibility when it is applied to environments with different road structures and geometries.

\subsection{Evaluation Scenario Configuration}

To approach and assess our evaluation criteria, we need a standardized process for comparing a simulated pedestrian generated by our model against a real-world pedestrian from a naturalistic data set. We devise a process to create \textit{evaluation scenarios}. These generated scenarios assist in validating our model against real-world data and provide a standard process that can be applied to any data set in a compatible format.

The idea of an evaluation scenario is to replace a single pedestrian in a given traffic recording with a simulated pedestrian and observe how it interacts with other pedestrians and whether it follows the same trajectory as the empirical pedestrian it replaced. In terms of a GeoScenario scenario, one pedestrian is selected as the \textit{evaluation pedestrian} and is created as an SP agent while all other pedestrians and vehicles in the recording are created as EP and EV agents respectively, and follow their corresponding trajectories from the data set. We refer to the data set pedestrian that the SP agent replaced as the \textit{empirical pedestrian} or alternatively, the evaluation pedestrian's \textit{empirical counterpart}. The scenario begins when the evaluation pedestrian enters the recording and ends when they exit.

The SP agent is initialized with three pieces of knowledge about its empirical counterpart: its starting position, its last position, and its average walking speed. In terms of the SFM component of the model, the last position is set as the SP's destination point and the average walking speed is set as the SP's desired speed.

The evaluation scenario creation process is repeated for each individual pedestrian in the data set, resulting in a unique evaluation scenario for each real-world pedestrian. As a result, each pedestrian in the data set produces their own evaluation scenario in which they are replaced by a model pedestrian that dynamically interacts with the other agents. This process compiles a suite of evaluation scenarios on which we can perform analysis and draw conclusions in terms of our evaluation criteria.

Two separate naturalistic data sets from different locations in Ontario, Canada, were used in the evaluation process.\footnote{\url{https://wiselab.uwaterloo.ca/waterloo-multi-agent-traffic-dataset/}} Each data set contains video files recorded by an overhead drone. The video files were then analyzed and relevant information, such as road user trajectory tracking and traffic light timings, was extracted and saved into a database. The first data set, referred to as the \textit{intersection} data set, was recorded at a busy four-way intersection with four signalized pedestrian crosswalks and two additional unsignalized crosswalks, each one across a right-turn merge lane (or slip lane). The second, referred to as the \textit{single crosswalk} data set, contains a single unsignalized crosswalk across a two-way two-lane road at a university.

During evaluation, we noted that the scenarios at the intersection location had relatively longer durations with an average of 66.82 seconds. Due to the minimal knowledge about the empirical trajectory, a concern arose that small deviations in the trajectories early in the scenario may amplify and compound into large deviations further into the scenario. These large deviations may not be representative of the model's performance at each moment in time and may be consequences of previous error. To mitigate this, we introduce segmented scenarios, in which each full evaluation scenario is subdivided into multiple segmented scenarios. Each segmented scenario represents one section of the pedestrian's journey spanning, for example, a single crosswalk or a single segment of sidewalk. For each evaluation scenario from the intersection data set, one or more additional segmented evaluation scenarios are created and grouped separately.

\begin{figure*}[ht]
\captionsetup[subfigure]{justification=centering}

\begin{subfigure}{0.32\linewidth}
\includegraphics[trim=0 0cm 1.5cm 1.25cm,clip,width=\linewidth]{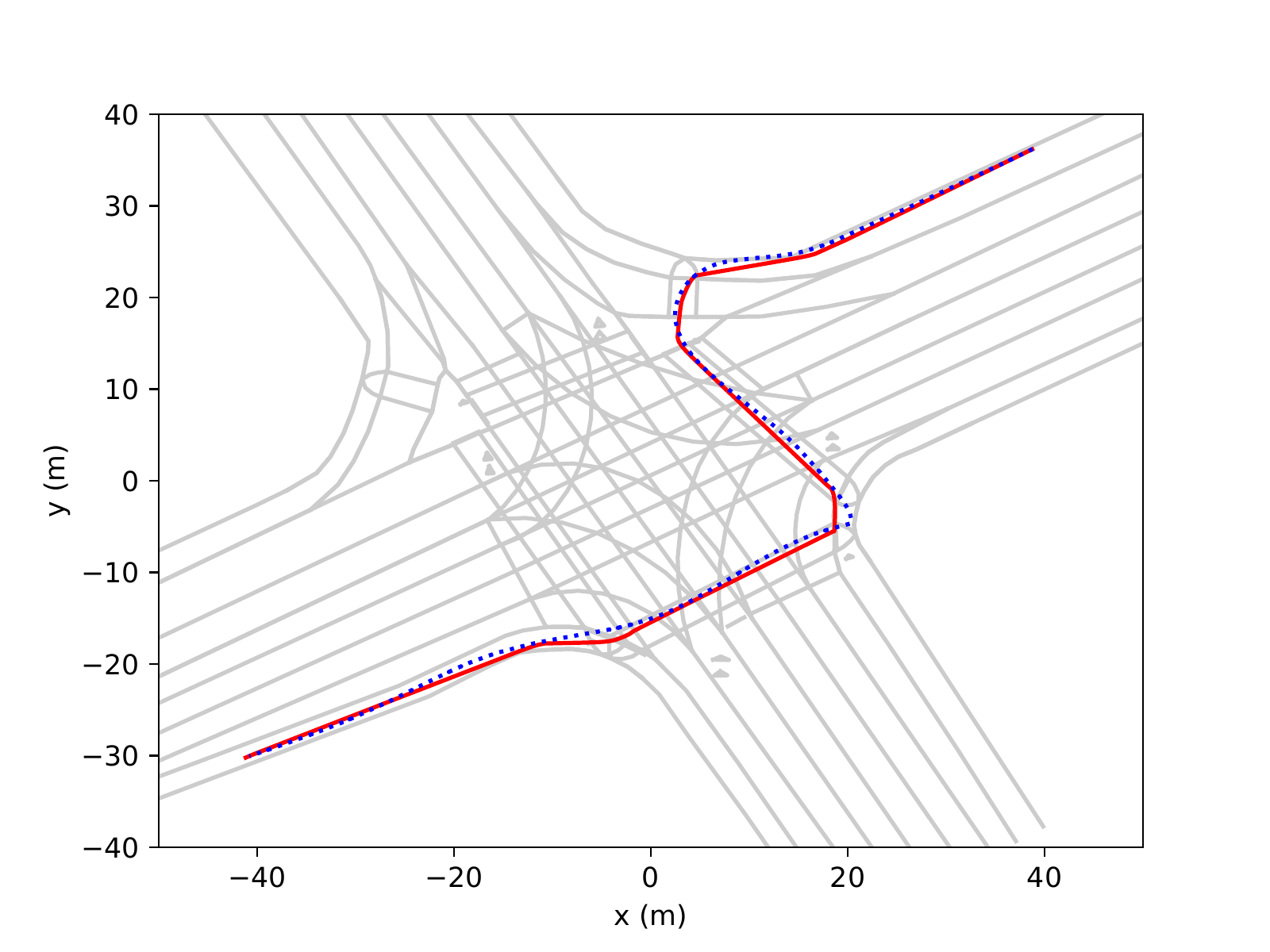}
\caption{Intersection location\\ED: 1.81\,m\\Fr\'{e}chet: 1.69\,m}
\label{subfig:uni_weber_ex_1}
\end{subfigure}
~
\begin{subfigure}{0.32\linewidth}
\includegraphics[trim=0 0cm 1.5cm 1.25cm,clip,width=\linewidth]{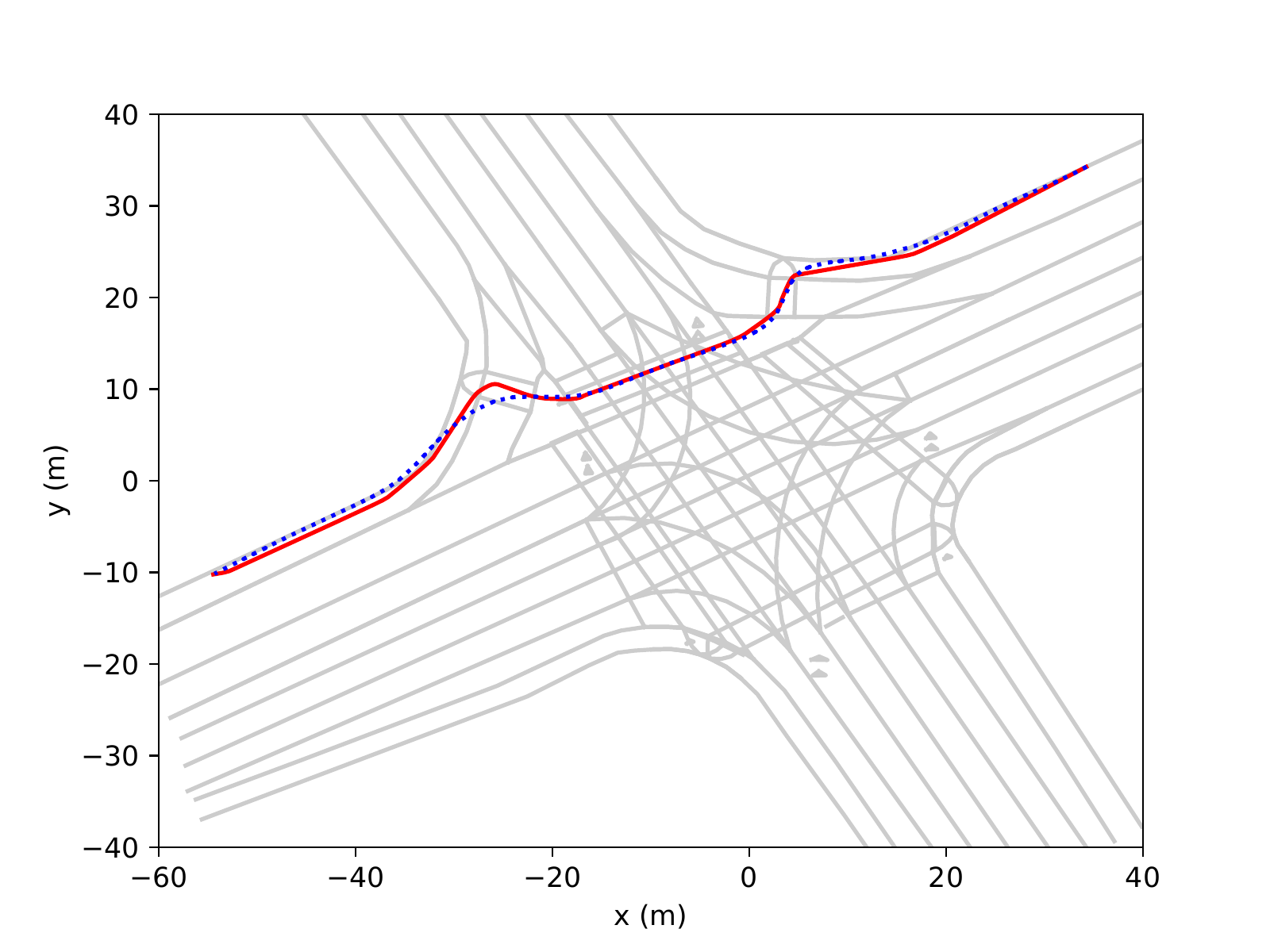}
\caption{Intersection location\\ED: 1.81\,m\\Fr\'{e}chet: 1.82\,m}
\label{subfig:uni_weber_ex_2}
\end{subfigure}
~
\begin{subfigure}{0.32\linewidth}
\includegraphics[trim=0 0cm 1.5cm 1.25cm,clip,width=\linewidth]{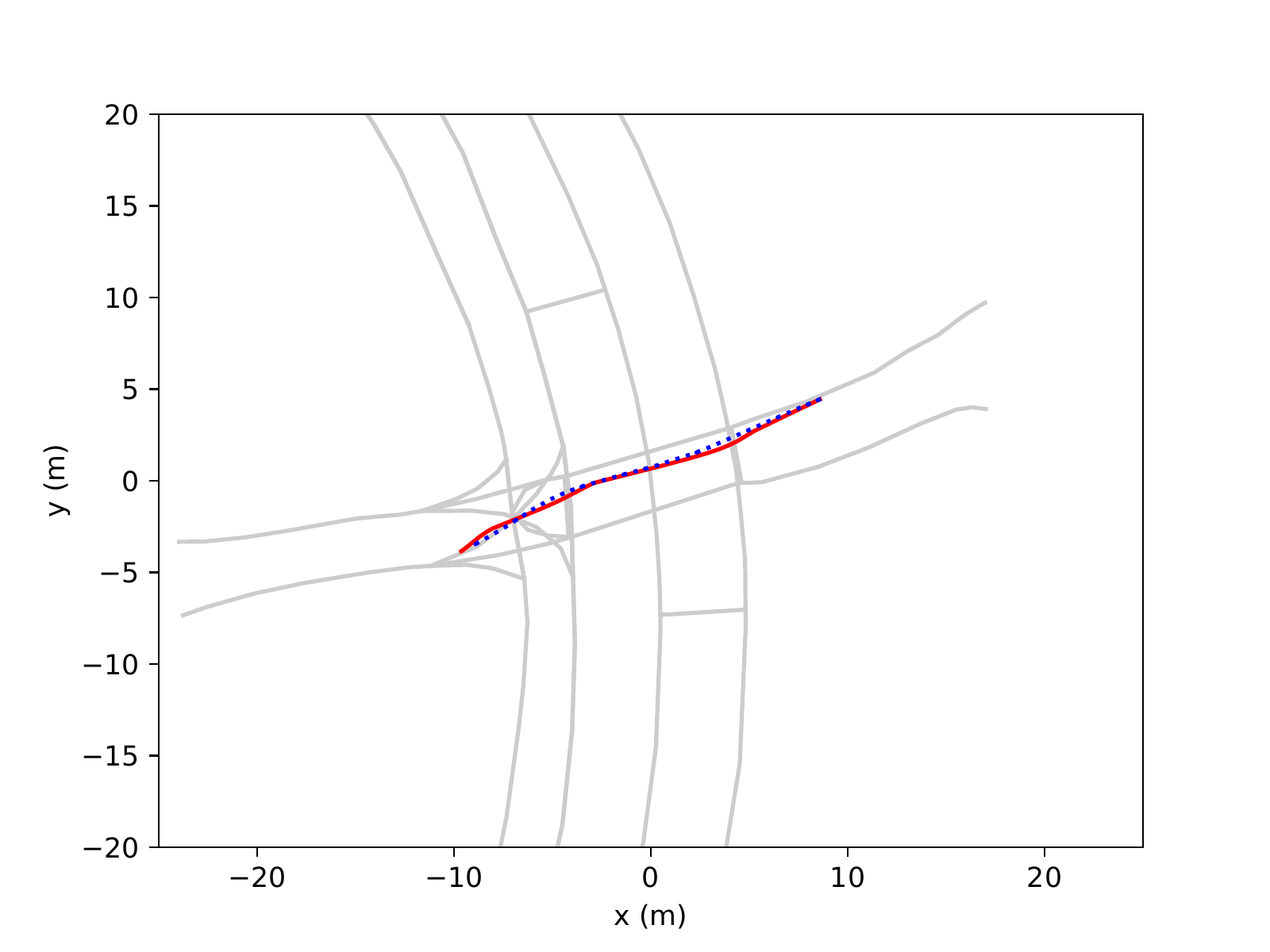}
\caption{Single crosswalk location\\ED: 0.14\,m\\Fr\'{e}chet: 0.32\,m}
\label{subfig:ring_road_ex_1}
\end{subfigure}
\caption{Trajectory traces of example evaluation scenarios with varying paths from both data set locations. Empirical trajectories are displayed as blue dots and the simulated trajectories are shown as a solid red line.}
\label{fig:traj_trace_examples}
\end{figure*}

\subsection{Realism of Low-Level Trajectories} \label{subsec:trajectory_metrics}

It is crucial that our behavior model produces pedestrian movements that are as natural and human-like as possible in order to be relied upon as a realistic representation of pedestrians in test scenarios. To evaluate our model's effectiveness at generating realistic motion, we run the evaluation scenarios and record each evaluation pedestrian's generated trajectory. We compare this simulated trajectory generated by our model with the trajectory of the corresponding empirical pedestrian by two trajectory-matching metrics: Euclidean distance and discrete Fr\'{e}chet distance. Given the dynamic nature of humans, it is unreasonable to expect a single configuration of our model to completely cover the varying behaviors of all the pedestrians in the data set. To accommodate this, we manually search for a custom configuration of parameter values for each evaluation scenario that best matches the corresponding real-world pedestrian's actions and behaviors.

The Euclidean distance (ED) measures the average distance between each pair of corresponding points between the simulated and empirical trajectories. The ED metric we use is also known as the spatio-temporal Euclidean distance (STED)~\cite{Nanni2006}. The points of each trajectory are equi-timed and recorded at each simulation time step in the evaluation scenario. On the other hand, the Fr\'{e}chet distance provides a measure of geometric similarity between the trajectories or paths. It does not depend on the velocity profiles as the Euclidean distance does and instead judges the similarity in ``shape'' of the two compared trajectories.

The majority of test cases during development of our model were derived from the intersection data set. While this data set provides a substantial number of pedestrians exhibiting a wide range of behaviors, a goal of any useful pedestrian model should be extensibility to differing road structures and geometries. We test the generalization and extensibility of our model by extracting evaluation scenarios from the second data set (single crosswalk). We evaluate the model's ability to navigate this new road structure with the same trajectory-matching metrics. Table~\ref{tab:trajectory_results} displays the average ED, the maximum ED, and the average FD over the scenarios on each data set location.

To give context to our model's results, we introduce a method to generate a baseline trajectory-approximation for each empirical pedestrian. We produce a baseline trajectory for each evaluation scenario based on the empirical path. We first define a set of points on the intersection map that can be connected in various orders to approximate the empirical paths. The selected map points closely resemble the waypoints a model pedestrian may select to navigate through the intersection. A baseline trajectory is constructed by manually selecting an ordered set of points, beginning and ending with the start and end points, respectively, that best approximates the data set pedestrian's path. Equi-distanced and equi-timed points are then linearly interpolated between these points to form a trajectory with the same number of points as the corresponding empirical trajectory.

\begin{table}[ht]
    \centering
    \caption{Trajectory-matching metrics}
    \label{tab:trajectory_results}
    \begin{tabular}{r | r | c c c c}
        \toprule
        \multirow{2}{*}{Location} & \multirow{2}{*}{Length} & \multirow{2}{*}{\# Scenarios} & ED & Max.\ ED & Fr\'{e}chet \\
        & & & (m) & (m) & (m) \\
        \midrule
        \multirow{3}{*}{Intersection} & Baseline & 198 & 5.45 & 12.48 & 2.90 \\
        & Full & 198 & 2.23 & 5.21 & 3.04 \\
        & Seg. & 527 & 1.55 & 2.65 & 1.81 \\
        \midrule
        Crosswalk & Full & 1017 & 1.36 & 2.01 & 1.71 \\
        \bottomrule
    \end{tabular}
\end{table}

It is interesting to note that the baseline method has a slightly improved Fr\'{e}chet distance over the model's full intersection scenarios. A reasonable explanation for this is that the baseline does not react to other pedestrians along the path, as it is simply a set of distances between waypoints. The model, on the other hand, interacts with other pedestrians and vehicles in their vicinity as do the empirical pedestrians. This introduces the risk of the model choosing a different course of action for their interaction than their empirical counterpart, for example, avoiding oncoming pedestrians on the right side instead of left, resulting in larger points of error. The baseline method, being non-reactive, averages out these deviations and may result in an improved geometric-based metric over a long duration scenario.

To visually confirm our model's trajectory-matching effectiveness, we trace both the simulated and empirical pedestrians' trajectories overlaid on a outline of the map file. \Cref{fig:traj_trace_examples} shows traces of evaluation scenarios at both the intersection location (\Cref{subfig:uni_weber_ex_1,subfig:uni_weber_ex_2}) and the single crosswalk location (\Cref{subfig:ring_road_ex_1}).

We note that the individual layers of the model are not evaluated independently. The Social Force model and its variations have been shown to effectively model both the individual movements of agents in sparse groups~\cite{Lakoba2005} and the crowd dynamics of dense groups~\cite{Helbing1995} \cite{Helbing2000} \cite{Mehran2009}. We focus on the conjunctive multi-layer use of high and low-level planners to produce realistic motion.

\subsection{Realism of High-Level Decisions}

For a pedestrian behavior model to properly represent the behaviors of real-world pedestrians, not only must it model the low-level trajectory movements, but it must also be able to replicate high-level decision-making processes. In order to measure this criterion, we need to define a metric by which we can conclusively declare that a decision made by a model pedestrian is the same decision made by its empirical counterpart. First, we must define a list of decisions to be observed from the naturalistic data. It was determined that there are two notable decisions pedestrians make while navigating an intersection.

\begin{enumerate}
    \item Given a set of accessible crosswalks and the requirement that at least one crosswalk must be crossed to reach the destination, which crosswalk is selected to cross
    \item Given a target crosswalk that the pedestrian has already decided to take which displays a red/yellow crossing light state, will the pedestrian begin to cross or wait until the next green state
\end{enumerate}

The above decisions were designed to ensure a binary response can be recorded for each instance of the decision across evaluation scenarios. The model pedestrian either makes the same decision as the data set pedestrian or they do not. In the rest of this section, we refer to these decisions as Decision 1 and 2 respectively. Note that this metric is only considered on the intersection data set due to the overly simplified road structure of the second data set location. For each full evaluation scenario, we noted the decision points where one of the two listed decisions were made. If the model pedestrian makes the same decision, a \textit{Same} data point is recorded, otherwise we record a \textit{Not Same} data point. \Cref{tab:decision_metric_results} displays the results of this process.

\begin{table}[ht]
    \caption{Summary of decision-based metric results on full length evaluation scenarios using the intersection data set}
    \label{tab:decision_metric_results}
    \centering
    \begin{tabular}{ r | c c c }
        \toprule
        \multirow{2}{*}{Decision} & Decision Points & \multicolumn{2}{c}{Same Decisions}\\
         & \# & \# & \% \\
        \midrule
        1 & 253 & 253 & 100.0 \\
        2 & 148 & 145 & 98.0 \\
        \bottomrule
    \end{tabular}
\end{table}

We see that our model pedestrians selected the same crosswalk when presented with multiple options as the real-world pedestrian 100\% of the time. However, there are three instances where the model pedestrian failed to enter the crosswalk at the same time as defined by Decision 2. After investigation, all three of these instances were due to the real-world pedestrian waiting at a red signal for most of its duration, then, when the vehicles' traffic lights show a two-way advanced green signal, they decide to enter the crosswalk as soon as there are no more conflicting left-turning vehicles. An adjustment to the High level of aggressiveness can account for this specific situation. A range of behaviors can be implemented, including a behavior where the probability of entering the crosswalk on a red signal increases with waiting time.

\subsection{Vehicle-Pedestrian Collision Scenario}

To showcase our model's practicality in creating critical scenarios, we revisit the scenario outlined in~\Cref{sec:intro}. The goal of this scenario is to demonstrate a plausible yet dangerous scenario that cannot be replicated with simple pedestrians that follow a constant speed profile. We created a scenario with one vehicle agent and one pedestrian agent in which the vehicle wants to pass through an unsignalized crosswalk that the pedestrian wants to cross. The pedestrian dynamically adjusts their walking or running speed to ensure a collision occurs at the crosswalk. At a technical level, this is achieved with our model by including a parameter in the \textbf{Keep in Lane} maneuver: \textit{collision\_vehicle = [vehicle\_id]}. Provided that the \textit{vehicle\_id} exists in the scenario and there is a crosswalk at which the agents can meet, the pedestrian dynamically determines a collision point based on the point of intersection between the crosswalk and the vehicle's heading vector. With assumed perfect perception of the vehicle's distance and speed, the pedestrian adjusts their own speed to ensure they reach the collision point at the same time as the vehicle.

A key benefit to using our model for this scenario is that the pedestrian will ensure a collision occurs regardless of the starting positions and velocities of the agents. With preset speed-profile pedestrians, the tester would need to manually reset the scenario's parameters with either trial and error or complex search-based computations to achieve the same result. Instead, with our approach, no changes need to be applied in order to force a collision scenario at every run.

\section{Discussion and Limitations} \label{sec:limitations}

In the current version of our model, we do not consider the implicit or explicit communication between human drivers and pedestrians, such as changes in speed or hand gestures. Such interactions are important considerations in modelling pedestrian movement and behavior. In our evaluation, we do not have access to this communication due to the data sets being recorded by overhead drones. Previous works, such as TrafficSim \cite{Suo2021}, have learned pedestrian behavior taking into account all forms of communication and external factors through data-driven approaches. Given the black-box nature of such approaches, we differentiate our work by providing finer control over individual pedestrian behaviors and interactions.

An extension of our model could incorporate human-driver interactions through behavior trees, specifically through more complex condition nodes. As long as the communication can be perceived by each agent, the pedestrian can evaluate the interaction and choose an appropriate action accordingly.

Adding to the set of available maneuvers and conditions can enhance the complexity and accuracy of the pedestrian's behavior. However, more maneuvers and conditions may also lead to a higher risk of incorrect or dangerous decisions. We must mediate the trade off between a manageable set of behaviors and their coverage and accuracy of real-world decisions, with evaluation on the available real-world data. Engineers must also be cautious of overfitting the behavior trees when designing critical scenarios not found in data. However, some scenario-based testing applications may require forcing a specific behavior to create a desired test scenario. In these cases, it is not so important to construct generalized behavior trees and it is sufficient to simply create the tree that generates the isolated behavior.

Since behavior trees allow for any number of parameters within their maneuver and condition nodes, it may be desirable to automate the process of tuning these parameters on real-world data. With separate sets of scenarios for tuning the trees and testing the model, the accuracy and robustness of the behavior trees can be improved. Using real-world data, the tuning process can also introduce the influence of implicit communication methods not previously incorporated into the model.

Another limitation of our model is the point-mass representation of pedestrians. The SP model abstracts characteristics from the agents, such as demographic information (age, gender, etc.) and body pose, which may be relevant to decisions for both drivers and pedestrians. For example, an elderly person will tend to move at different speeds and make different crossing decisions to an adult or a child in the same situation. We anticipate more sophisticated representations of pedestrians in future iterations of our model.

Finally, our model does not consider the uncertainty and noise of human motion. This is especially relevant in interactions between pedestrians and autonomous vehicles. Though not included in this iteration of the SP model, uncertainty can be injected into behavior trees. Noise can be used in condition nodes to express perception or judgement errors made by humans and in maneuver nodes to produce unexpected movements or decisions that may be difficult for an AV to predict.
\section{Conclusions} \label{sec:conclusions}

In this paper, we presented a novel hierarchical pedestrian behavior model that is capable of producing realistic trajectories through different traffic environments while following the rules of the road and making rational real-time decisions, but also allowing for misbehaviors. A multi-layer approach was applied to this problem that incorporates a high-level Behavior layer that determines an appropriate maneuver to be processed by the Maneuver layer, which informs the Motion Planner layer on how to adjust the low-level trajectory movements. The conjunctive use of behavior trees with an adapted Social Force Model ensures the model's agents are an accurate representation of real-world pedestrians. They were shown to make the same decisions when faced with multiple options and also display natural movements and interactions with other pedestrians and vehicles in the scene.

Our presented model offers benefit to the scenario-based testing of autonomous vehicles. Since pedestrian decisions and actions can be explicitly represented with behavior trees, engineers are able to inject desired behaviors into scenarios to test the AV's responses to critical situations. We provide an implementation of a set of basic maneuvers and conditions shown to sufficiently cover real-world behaviors. The flexibility and modularity of behavior trees allow for extensions of this list and configurations of trees that cover a wide range of conceivable behaviors.

We evaluated our model in terms of its ability to produce realistic low-level movements in two environments with different road structures and to replicate high-level decisions made by real-world pedestrians observed in a naturalistic data set. The results of our evaluation confirm that we present a viable pedestrian simulation model capable of producing realistic pedestrian decisions and movements.

\addtolength{\textheight}{-8.1cm}   



\IEEEtriggeratref{10}
\IEEEtriggercmd{\balance}

\bibliographystyle{IEEEtran}

\bibliography{references}

\begin{thebibliography}{10}
\providecommand{\url}[1]{#1}
\csname url@samestyle\endcsname
\providecommand{\newblock}{\relax}
\providecommand{\bibinfo}[2]{#2}
\providecommand{\BIBentrySTDinterwordspacing}{\spaceskip=0pt\relax}
\providecommand{\BIBentryALTinterwordstretchfactor}{4}
\providecommand{\BIBentryALTinterwordspacing}{\spaceskip=\fontdimen2\font plus
\BIBentryALTinterwordstretchfactor\fontdimen3\font minus
  \fontdimen4\font\relax}
\providecommand{\BIBforeignlanguage}[2]{{%
\expandafter\ifx\csname l@#1\endcsname\relax
\typeout{** WARNING: IEEEtran.bst: No hyphenation pattern has been}%
\typeout{** loaded for the language `#1'. Using the pattern for}%
\typeout{** the default language instead.}%
\else
\language=\csname l@#1\endcsname
\fi
#2}}
\providecommand{\BIBdecl}{\relax}
\BIBdecl

\bibitem{Kentaro2004}
K.~Go and J.~M. Carroll, ``The blind men and the elephant: Views of
  scenario-based system design,'' \emph{Interactions}, vol.~11, no.~6, pp.
  44--53, Nov 2004.

\bibitem{Millard-Ball2018}
A.~Millard-Ball, ``Pedestrians, autonomous vehicles, and cities,''
  \emph{Journal of Planning Education and Research}, vol.~38, no.~1, pp. 6--12,
  2018.

\bibitem{CamaraFanta2018}
F.~Camara, O.~Giles, R.~Madigan, M.~Rothmüller, P.~H. Rasmussen, S.~A.
  Vendelbo-Larsen, G.~Markkula, Y.~M. Lee, L.~Garach, N.~Merat, and C.~W. Fox,
  ``Predicting pedestrian road-crossing assertiveness for autonomous vehicle
  control,'' in \emph{2018 21st International Conference on Intelligent
  Transportation Systems (ITSC)}, 2018, pp. 2098--2103.

\bibitem{LiChuanyao2021}
C.~Li, S.~Liu, and X.~Cen, ``Safety and efficiency impact of
  pedestrian–vehicle conflicts at non signalized midblock crosswalks based on
  fuzzy cellular automata,'' \emph{Physica A: Statistical Mechanics and its
  Applications}, vol. 572, p. 125871, 2021.

\bibitem{Li2012}
X.~Li, X.~Yan, X.~Li, and J.~Wang, ``Using cellular automata to investigate
  pedestrian conflicts with vehicles in crosswalk at signalized intersection,''
  \emph{Discrete Dynamics in Nature and Society}, vol. 2012, p. 287502, Nov
  2012.

\bibitem{Feng2013}
S.~Feng, N.~Ding, T.~Chen, and H.~Zhang, ``Simulation of pedestrian flow based
  on cellular automata: A case of pedestrian crossing street at section in
  {C}hina,'' \emph{Physica A: Statistical Mechanics and its Applications}, vol.
  392, no.~13, pp. 2847--2859, 2013.

\bibitem{Suo2021}
S.~Suo, S.~Regalado, S.~Casas, and R.~Urtasun, ``Trafficsim: Learning to
  simulate realistic multi-agent behaviors,'' in \emph{Proceedings of the
  IEEE/CVF Conference on Computer Vision and Pattern Recognition (CVPR)}, June
  2021, pp. 10\,400--10\,409.

\bibitem{Helbing1998}
D.~Helbing and P.~Molnar, ``Self-organization phenomena in pedestrian crowds,''
  arXiv:\,cond-mat/9806152, 1998.

\bibitem{Blue2001}
V.~J. Blue and J.~L. Adler, ``Cellular automata microsimulation for modeling
  bi-directional pedestrian walkways,'' \emph{Transportation Research Part B:
  Methodological}, vol.~35, no.~3, pp. 293--312, 2001.

\bibitem{Zhang2012}
P.~Zhang, X.-X. Jian, S.~C. Wong, and K.~Choi, ``Potential field cellular
  automata model for pedestrian flow,'' \emph{Phys. Rev. E}, vol.~85, p.
  021119, Feb 2012.

\bibitem{Anvari2015}
B.~Anvari, M.~G. Bell, A.~Sivakumar, and W.~Y. Ochieng, ``Modelling shared
  space users via rule-based social force model,'' \emph{Transportation
  Research Part C: Emerging Technologies}, vol.~51, pp. 83--103, 2015.

\bibitem{Rinke2017}
N.~Rinke, C.~Schiermeyer, F.~Pascucci, V.~Berkhahn, and B.~Friedrich, ``A
  multi-layer social force approach to model interactions in shared spaces
  using collision prediction,'' \emph{Transportation Research Procedia},
  vol.~25, pp. 1249--1267, 2017, world Conference on Transport Research - WCTR
  2016 Shanghai. 10-15 July 2016.

\bibitem{Tanimoto2010}
J.~Tanimoto, A.~Hagishima, and Y.~Tanaka, ``Study of bottleneck effect at an
  emergency evacuation exit using cellular automata model, mean field
  approximation analysis, and game theory,'' \emph{Physica A: Statistical
  Mechanics and its Applications}, vol. 389, no.~24, pp. 5611--5618, 2010.

\bibitem{Guan2016}
J.~Guan, K.~Wang, and F.~Chen, ``A cellular automaton model for evacuation flow
  using game theory,'' \emph{Physica A: Statistical Mechanics and its
  Applications}, vol. 461, pp. 655--661, 2016.

\bibitem{Lin2018}
G.-W. Lin and S.-K. Wong, ``Evacuation simulation with consideration of
  obstacle removal and using game theory,'' \emph{Phys. Rev. E}, vol.~97, p.
  062303, Jun 2018.

\bibitem{Wu2019}
W.~Wu, R.~Chen, H.~Jia, Y.~Li, and Z.~Liang, ``Game theory modeling for
  vehicle–pedestrian interactions and simulation based on cellular
  automata,'' \emph{International Journal of Modern Physics C}, vol.~30,
  no.~04, p. 1950025, 2019.

\bibitem{Helbing2000}
D.~Helbing, I.~Farkas, and T.~Vicsek, ``Simulating dynamical features of escape
  panic,'' \emph{Nature}, vol. 407, no. 6803, pp. 487--490, Sep 2000.

\bibitem{Colledanchise2018}
M.~Colledanchise and P.~{\"O}gren, \emph{Behavior Trees in Robotics and {AI}:
  An Introduction}, 1st~ed., ser. Artificial Intelligence and Robotics.\hskip
  1em plus 0.5em minus 0.4em\relax Taylor \& Francis CRC Press, 2018.

\bibitem{modes2020-01-1204}
M.~Antkiewicz, M.~Kahn, M.~Ala, K.~Czarnecki, P.~Wells, A.~Acharya, and
  S.~Beiker, ``Modes of automated driving system scenario testing: Experience
  report and recommendations,'' \emph{SAE International Journal of Advances and
  Current Practices in Mobility}, vol.~2, no.~4, pp. 2248--2266, apr 2020.

\bibitem{Queiroz2019}
R.~Queiroz, T.~Berger, and K.~Czarnecki, ``{Geo}{S}cenario: An open {DSL} for
  autonomous driving scenario representation,'' in \emph{IEEE Intelligent
  Vehicles Symposium (IV)}, IEEE.\hskip 1em plus 0.5em minus 0.4em\relax Paris:
  IEEE, 2019.

\bibitem{Poggenhans2018}
F.~Poggenhans, J.-H. Pauls, J.~Janosovits, S.~Orf, M.~Naumann, F.~Kuhnt, and
  M.~Mayr, ``{L}anelet2: A high-definition map framework for the future of
  automated driving,'' in \emph{2018 21st International Conference on
  Intelligent Transportation Systems (ITSC)}, 2018, pp. 1672--1679.

\bibitem{Queiroz2022_arXiv}
R.~Queiroz, D.~Sharma, R.~Caldas, K.~Czarnecki, S.~Garc\'{i}a, T.~Berger, and
  P.~Pelliccione, ``A driver-vehicle model for {ADS} scenario-based testing,''
  arXiv:\,2205.02911, 2022.

\bibitem{CARLA-pmlr-v78-dosovitskiy17a}
A.~Dosovitskiy, G.~Ros, F.~Codevilla, A.~Lopez, and V.~Koltun, ``{CARLA}: {An}
  open urban driving simulator,'' in \emph{Proceedings of the 1st Annual
  Conference on Robot Learning}, ser. Proceedings of Machine Learning Research,
  S.~Levine, V.~Vanhoucke, and K.~Goldberg, Eds., vol.~78.\hskip 1em plus 0.5em
  minus 0.4em\relax PMLR, 13--15 Nov 2017, pp. 1--16.

\bibitem{Nanni2006}
M.~Nanni and D.~Pedreschi, ``Time-focused clustering of trajectories of moving
  objects,'' \emph{J. Intell. Inf. Syst.}, vol.~27, pp. 267--289, 11 2006.

\bibitem{Lakoba2005}
T.~I. Lakoba, D.~J. Kaup, and N.~M. Finkelstein, ``Modifications of the
  {Helbing}-{Molnar}-{Farkas}-{Vicsek} social force model for pedestrian
  evolution,'' \emph{Simulation}, vol.~81, no.~5, pp. 339--352, 2005.

\bibitem{Helbing1995}
D.~Helbing and P.~Moln\'ar, ``Social force model for pedestrian dynamics,''
  \emph{Phys. Rev. E}, vol.~51, pp. 4282--4286, May 1995.

\bibitem{Mehran2009}
R.~Mehran, A.~Oyama, and M.~Shah, ``Abnormal crowd behavior detection using
  social force model,'' in \emph{2009 IEEE Conference on Computer Vision and
  Pattern Recognition}, 2009, pp. 935--942.

\end{thebibliography}


\end{document}